\documentclass[runningheads]{llncs}
\usepackage[T1]{fontenc}
\usepackage{graphicx,verbatim}
\usepackage{amsmath,amssymb}
\usepackage{booktabs}
\usepackage{multirow}
\usepackage{xcolor}
\usepackage{url}
\usepackage{color}
\usepackage{hyperref}
\usepackage{marvosym}

\begin{document}
\title{CDPM-Align: Multi-Scale Guidance-Aligned Diffusion Pretraining for Robust Few-Shot Anatomical Landmark Detection}
\titlerunning{CDPM-Align for Robust Landmark Detection}
\author{Roberto Di Via\inst{1}\orcidID{0009-0005-8907-8385} \and
Irina Voiculescu\inst{2}\orcidID{0000-0002-9104-8012} \and
Francesca Odone\inst{1}\orcidID{0000-0002-3463-2263} \and
Vito Paolo Pastore\inst{1}\orcidID{0000-0002-5827-5571}\textsuperscript{(\Letter)}}

\authorrunning{R. Di Via et al.}

\institute{MaLGa, DIBRIS, University of Genoa, Italy\\
\Letter\ \email{vito.paolo.pastore@unige.it} \and
Department of Computer Science, University of Oxford, UK}

\maketitle              
\begin{abstract}
Anatomical landmark detection is a fundamental task in medical image analysis supporting a wide range of diagnostic and interventional workflows. Although recent methods have achieved sub-millimetric localisation, accuracy alone is not sufficient for clinical deployment, requiring reliability and robustness in prediction. 
Despite its clinical relevance, the impact of representation learning in this context is still underexplored. 
In this work, we introduce CDPM-align, a multi-scale guidance-aligned conditional diffusion pre-training for anatomical landmark detection. Our experimental setup focuses on a few images and a few annotation regimes. Specifically, we employ three popular heterogeneous small-scale benchmark datasets for representation learning via conditional generative pre-training. Furthermore, we consider low-annotation scenarios for the downstream task of landmark detection, with 10 and 25 annotated images, reflecting realistic trade-offs between clinical effort and resource constraints for annotations. Our results confirm that generative pre-training enables the model to learn a robust representation. This improves both accuracy and uncertainty on the downstream tasks, advancing towards safe and efficient clinical deployment.


\end{abstract}
\section{Introduction}
\label{sec:intro}

Anatomical landmark detection in medical X-ray images underpins cephalometric treatment planning~\cite{wang2016benchmark}, cardiothoracic ratio estimation~\cite{antromonica2025deep}, and skeletal maturity assessment~\cite{li2023bone}. 
Although modern deep learning methods can achieve sub-millimetre localisation accuracy~\cite{payer2019integrating,stansfield2025landmark,huang2025h3denet,zhou2024hybrid,wyatt2024optimising}, point-wise error alone does not guarantee clinical reliability~\cite{CleAll_Confidence_MICCAI2025,divia2025xray,divia2026hip}.
A clinically trustworthy model must not only predict an accurate coordinate, but also concentrate probability mass tightly around the true landmark, reflecting calibrated spatial uncertainty. 
To quantify this property, the Expected Radial Error (ERE) was introduced for anatomical landmark detection~\cite{mccouat2022contour}. ERE measures the expected Euclidean distance between the predicted landmark and a sample from the predicted heatmap, jointly capturing localisation accuracy and distributional sharpness, providing a more faithful assessment of reliability. For instance, a model with accurate mean predictions but diffuse heatmaps would incur a high ERE, indicating decreased reliability. Despite its clinical relevance, how pretraining influences ERE and robustness, particularly under limited annotation budgets, remains underexplored.
Traditionally, medical landmark detection has relied on supervised pretraining, either on large natural image datasets, such as ImageNet, or on domain-specific X-rays~\cite{patel2025handful,divia2024indomain}. However, these approaches often fail to capture nuances across different anatomical regions. Self-supervised learning (SSL) methods, including DINO~\cite{caron2021emerging}, SimCLR\,v2~\cite{chen2020big}, MoCo\,v3~\cite{chen2021mocov3}, and pixel-level embeddings such as SAM~\cite{Yan_2022_TMI}, learn transferable dense features without extensive labels~\cite{truong2021transferable}, yet their performance plateaus under extreme data scarcity and structural heterogeneity. Universal detectors~\cite{zhu2021you} and foundation models like MedSapiens~\cite{elbatel2025medsapiens} aggregate multiple datasets for broad generalisation but primarily optimise point-wise accuracy, neglecting local distributional sharpness and uncertainty.
Generative pretraining with diffusion models offers a complementary approach to learn spatially organised semantics suitable for dense prediction~\cite{baranchuk2022label}. Di~Via~et~al.~\cite{divia2025diffusion} showed unconditional diffusion can improve few-shot landmark detection, but without conditioning, anatomical modes are conflated and anatomical region-specific structural representations remain underconstrained. Across these methods, the effect of pretraining on predictive uncertainty and robustness has been unexplored, motivating approaches that enforce both global context and local structural consistency for reliable heatmap estimation.
\begin{figure}[t]
  \centering
  \includegraphics[width=\linewidth]{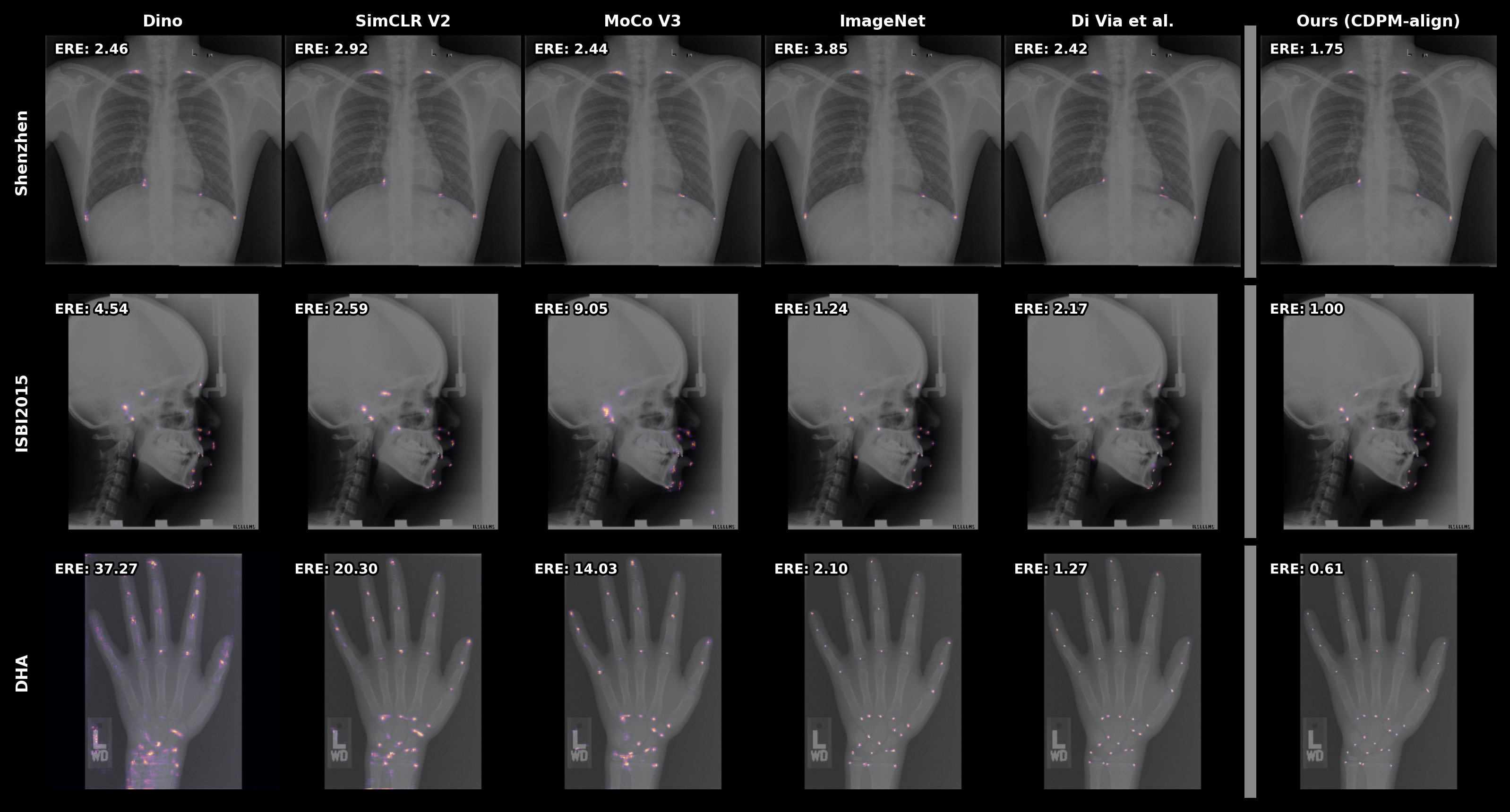}
    \caption{Predicted heatmap overlays at 10-shot across datasets (rows) and selected methods (columns). Our CDPM-align (rightmost) yields consistently lower uncertainty.}
  \label{fig:ere_comparison}
\end{figure}
Motivated by these limitations, we propose \textbf{CDPM-align}, a conditional diffusion pretraining framework tailored to few-shot anatomical landmark detection. Starting from small-scale heterogeneous collections of data, we condition diffusion training exploiting the specific dataset indices as labels, thus enabling the model to learn dataset-specific representations. We exploit the classifier-free guidance signal as an explicit structural descriptor, and introduce a multi-scale alignment objective that enforces directional features consistency across independently sampled diffusion timesteps and UNet hierarchy levels. This constraint encourages class-conditional structure that is stable across noise levels while preserving spatial fidelity. The pretrained backbone is subsequently fine-tuned end-to-end for the anatomical landmark detection task, using as few as 10--25 annotated images per dataset. Our main contributions can be summarised as follows:
\begin{itemize}
    \item We introduce a conditional generative pretraining strategy for small-scale heterogeneous X-ray datasets, augmented with a multi-scale guidance alignment loss that improves localisation accuracy, uncertainty concentration, and robustness in low-shot regimes.

    \item We validate the proposed method across three clinical dimensions (accuracy, predictive uncertainty, and robustness) on three public benchmarks under 10- and 25-shot protocols, showing consistent improvements over supervised, SSL, and state-of-the-art baselines.
\end{itemize}

\section{Method}

\begin{figure}[t]
  \centering
  \includegraphics[width=0.9\linewidth]{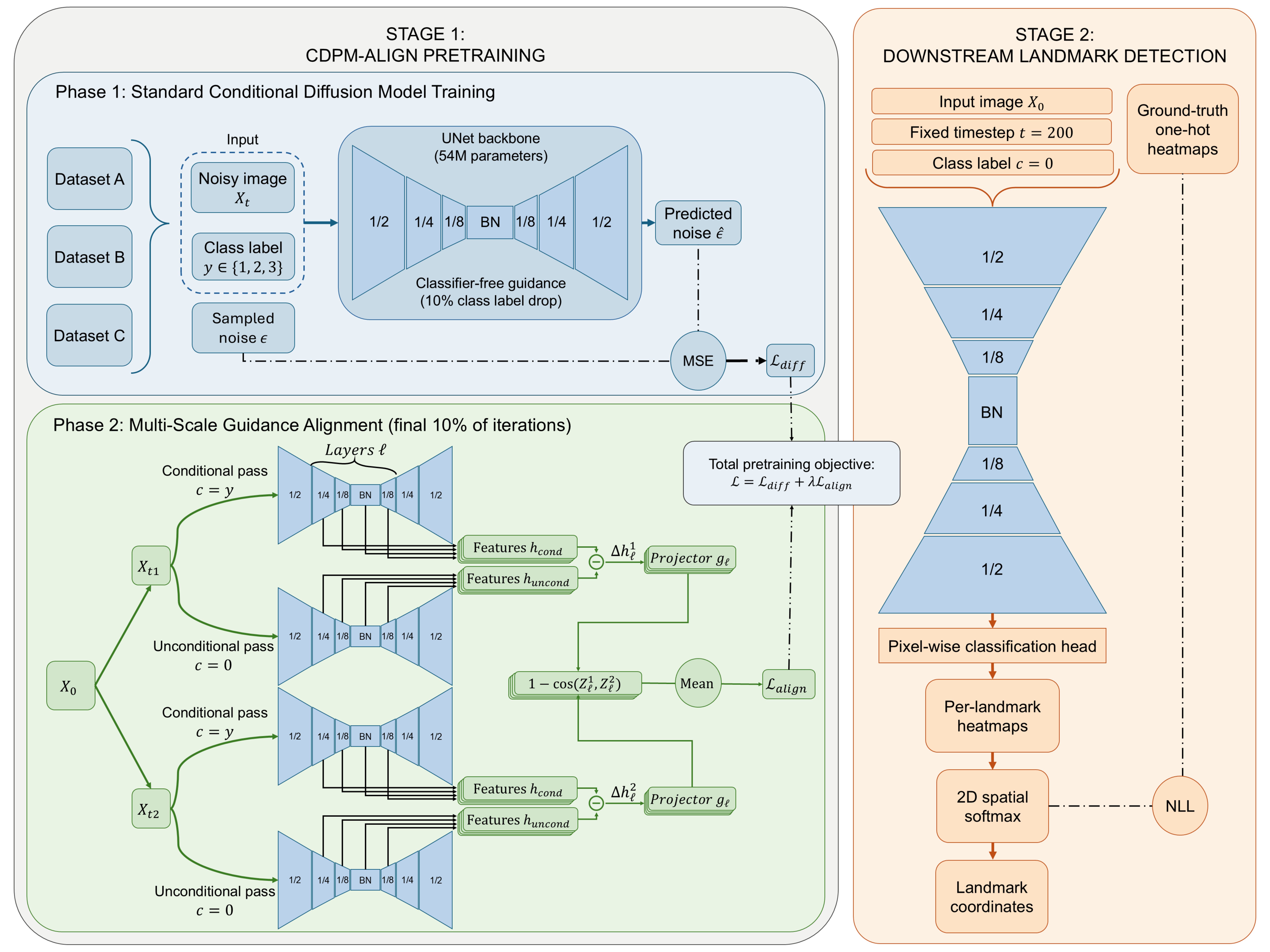}
  \caption{Overview of CDPM-align. \emph{Left}: Two-phase multi-dataset pretraining. Standard conditional diffusion training is followed by an alignment phase (final 10\% of iterations) enforcing directional consistency of the guidance signal across timesteps at four UNet scales. \emph{Right}: End-to-end few-shot fine-tuning via pixel-wise classification.}
  \label{fig:pipeline}
\end{figure}

CDPM-align consists of a pretraining stage followed by downstream adaptation (Sec. \ref{sec:downstream}). The pretraining stage itself has two phases (Sec. \ref{sec:cdpm}--\ref{sec:align}): the first trains a conditional diffusion probabilistic model (CDPM) with the standard generative objective; the second fine-tunes the same model for a short additional phase with the guidance alignment loss activated. The pretrained backbone is then adapted for few-shot anatomical landmark detection. A schematic overview of CDPM-align is provided in Fig.~\ref{fig:pipeline}.

\subsection{Conditional Diffusion Probabilistic Model (CDPM)}
\label{sec:cdpm}

We build on denoising diffusion probabilistic models (DDPMs)~\cite{ho2020denoising} with class-conditional generation via classifier-free guidance~\cite{ho2022classifier}. The forward process produces a noisy sample $x_t = \sqrt{\bar{\alpha}_t}\,x_0 + \sqrt{1-\bar{\alpha}_t}\,\epsilon$ with $\epsilon \sim \mathcal{N}(0,I)$, where $\bar{\alpha}_t = \prod_{s=1}^{t}(1-\beta_s)$, linear schedule $\beta_1{=}10^{-4}$ to $\beta_T{=}0.028$, $T{=}500$. A noise-prediction network $\epsilon_\theta(x_t, t, c)$ is trained with class label $c \in \{0,1,2,3\}$ (0 = unconditional; 1--3 = dataset index), with 10\% dropout to $c{=}0$ enabling CFG at inference, minimising $\mathcal{L}_{\text{diff}} = \mathbb{E}_{x_0,\epsilon,t}[\|\epsilon - \epsilon_\theta(x_t, t, c)\|^2]$. The backbone is a UNet~\cite{ronneberger2015u} with FiLM~\cite{perez2018film} embeddings for timestep and class conditioning.

\subsection{CDPM-Align: Multi-scale Alignment on the Guidance Signal}
\label{sec:align}
For each image $x_0$ belonging to dataset $y$, we sample two timesteps $t_1, t_2 \sim p(t)$, produce the noisy images $x_{t_1}$ and $ x_{t_2}$, and perform four UNet forward passes---$f_\theta(x_{t_i}, t_i, c)$ for $i \in \{1,2\}$ and $c \in \{y, \varnothing\}$---obtaining conditional $h_{\text{cond}}$ and unconditional $h_{\text{uncond}}$ features  at each timestep. 
The guidance difference $\Delta h = h_{\text{cond}} - h_{\text{uncond}}$ encodes dataset-specific anatomy; its magnitude varies with timestep while our intuition is that enforcing directional consistency across independently sampled timesteps encourages the UNet encoder to learn representations invariant to noise level.
We bias sampling toward mid-range timesteps $[T/4, \, 3T/4]$, since very early timesteps yield near-clean images where the guidance difference $\Delta h$ carries little class information, while very late timesteps are dominated by noise; the mid-range provides the richest dataset-discriminative signal. 
Features are extracted at four hierarchy levels:
\[\mathcal{S} = \{\text{encoder}_{1/4},\, \text{encoder}_{1/8},\, \text{bottleneck},\, \text{decoder}_{1/8}\}.\]
The per-layer guidance difference at level $\ell \in \mathcal{S}$ is $\Delta h^{(i)}_\ell = f_\theta(x_{t_i},t_i,y)[\ell] - f_\theta(x_{t_i},t_i,\varnothing)[\ell]$ for $i \in \{1,2\}$. 
At each level independently, $\Delta h^{(i)}_\ell$ is processed by a level-specific projection head $g_\ell$---comprising global average pooling (GAP), a two-layer MLP$_\ell$ ($C_\ell{\to}512{\to}256$, GELU), and $\ell_2$-normalisation---to yield $z^{(i)}_\ell = g_\ell(\Delta h^{(i)}_\ell)$.
The alignment loss averages the per-level cosine dissimilarity between the two views:
\[\mathcal{L}_{\text{align}} = \tfrac{1}{|\mathcal{S}|}\sum_{\ell \in \mathcal{S}}\!\bigl(1 - \cos(z^{(1)}_\ell, z^{(2)}_\ell)\bigr).\]
During the alignment phase, $\mathcal{L}_{\text{diff}}$ is computed over all four noise predictions (for timesteps ${t_1, t_2}$ and for conditions $c{=}0$, $c{=}y$), since the unconditional passes required for $\Delta h$ are already computed and supervising them preserves both generation pathways needed by CFG. The total objective is:
\[\mathcal{L} = \mathcal{L}_{\text{diff}} + \lambda_{\text{align}}\,\mathcal{L}_{\text{align}}.\]
The projection heads $g_\ell$ are discarded after pretraining; full UNet weights are transferred downstream.
Intuitively, our feature alignment is mostly effective when the diffusion model has learned to estimate the noise added to input images. Furthermore, it brings an additional computational overhead, as it requires to perform four forward passes. As such, we employ our alignment loss only when we reach the final $10\%$ of training iterations, when we fine-tune the CDPM model pre-trained in the first phase with $\mathcal{L} = \mathcal{L}_{\text{diff}}$. In this way, we manage to reduce the computational overhead from 4$\times$ to 1.3$\times$ with respect to naive training.

\subsection{Downstream Landmark Detection}
\label{sec:downstream}

Following~\cite{baranchuk2022label}, we use a fixed forward timestep ($t{=}200$) before feature extraction, yielding semantically organised intermediate representations. For anatomical landmark detection, the full pretrained UNet is fine-tuned end-to-end with a pixel-wise classification head, as in~\cite{mccouat2022contour}. 
During fine-tuning and inference, the unconditional class label ($c{=}0$) is used, decoupling the downstream task from pretraining labels, theoretically enabling future application to unseen datasets, while preventing over-reliance on the class token in few-shot settings.

\noindent\textit{Metrics.}
We report MRE (mean radial error), ERE (expected radial error, quantifying spatial uncertainty as the expected distance between a heatmap-sampled point and the predicted landmark), SDR@2 (successful detection rate within 2\,mm or 2\,px), and P95 (95th-percentile error, reflecting worst-case tail behaviour). Metrics are reported in millimetres, where possible (ISBI2015 and DHA, where the DICOM pixel spacing is available for pixel-millimetres conversion), and in pixels where such information is not available (Shenzhen). 

\section{Experiments}
\label{sec:experiments}

\subsection{Experimental Setup}
\label{sec:setup}

\noindent\textit{Datasets.}
We evaluate on three public benchmarks: \emph{Shenzhen}~\cite{jaeger2014two} (279 antero-posterior chest radiographs, 6 landmarks), \emph{ISBI2015}~\cite{wang2016benchmark} (400 lateral cephalograms, 19 landmarks), and the \emph{Digital Hand Atlas} (DHA)~\cite{gertych2007bone} (910 hand radiographs, 37 landmarks). All images are resized to $256\times256$ pixels preserving the original aspect ratio. The pooled unlabeled pretraining corpus, excluding the test set, comprises 988 images across all three datasets.
\noindent\textit{Baselines and few-shot protocol, and implementation.}
We evaluate under 10- and 25-shot annotation budgets, reporting mean\,$\pm$\,std over 5 independent runs. External baselines include: (i)~supervised UNet with ResNet-101 encoder (ImageNet pretrained); (ii)~three self-supervised UNet baselines~\cite{pyssl2023giakoumoglou} (DINO, MoCo\,v3, SimCLR\,v2) pretrained on the same pooled radiograph corpus with a ResNet-101 backbone; (iii)~unconditional diffusion pretraining~\cite{divia2025diffusion}; and (iv) a universal multi-anatomy detector~\cite{zhu2021you}. ResNet-101 is used to match CDPM’s ${\sim}$52M parameters and skip-connection capacity, as diffusion backbones cannot directly reuse discriminatively pretrained weights. Internal ablations include CDPM Scratch (random initialisation).
CDPM-align is pretrained for 50k iterations (45k standard diffusion, 5k alignment fine-tuning) with a batch size of 16 and the AdamW optimiser. Downstream fine-tuning runs for 200 epochs with early stopping, the NLL loss, and AdamW with an lr of $10^{-4}$ and a batch size of 8. Code will be released upon acceptance to support reproducibility.

\subsection{Main Results}
\label{sec:main_results}

Tables~\ref{tab:shenzhen}--\ref{tab:dha} compare all methods across datasets and annotation budgets. Both Friedman and Kruskal--Wallis statistical tests confirms global performance differences ($p \leq 0.001$), and CDPM-align achieves consistent superiority (5/5 runs) over all supervised, self-supervised, and state-of-the-art baselines~\cite{zhu2021you,divia2025diffusion}.

\paragraph{Accuracy.}

CDPM-align achieves the best MRE in 5 of 6 dataset–budget settings; the only exception is Shenzhen at 10-shot, where CDPM (NIH) benefits from larger-scale pretraining. Gains are substantial: +22\% on ISBI2015 10-shot (2.11 vs.\ 2.70\,mm) and +42\% on DHA 10-shot (2.51 vs.\ 4.34\,mm). At 25-shot, CDPM-align reaches 1.54\,mm on ISBI2015 (77.52\% SDR), approaching the 2\,mm clinical threshold with only 25 images.
On Shenzhen, CDPM-align achieves 3.01\,px (10-shot), outperforming the full-label method of~\cite{choi2025learning} (3.82\,px). Although the foundation model MedSapiens~\cite{elbatel2025medsapiens} reports 1.24\,mm (ISBI2015) and 3.72\,px (Shenzhen) with full supervision, our method is competitive using only 10–25 annotations.


\begin{table}[t]
\centering
\small
\caption{Landmark detection on \textbf{Shenzhen}. Best results per column are bolded.}
\label{tab:shenzhen}
\resizebox{\columnwidth}{!}{%
\begin{tabular}{lcccccccc}
\toprule
\multirow{2}{*}{\textbf{Method}}
& \multicolumn{4}{c}{\textbf{10-shot}}
& \multicolumn{4}{c}{\textbf{25-shot}} \\
\cmidrule(lr){2-5} \cmidrule(lr){6-9}
& \textbf{MRE}$\downarrow$ & \textbf{ERE}$\downarrow$ & \textbf{SDR@2}$\uparrow$ & \textbf{P95}$\downarrow$
& \textbf{MRE}$\downarrow$ & \textbf{ERE}$\downarrow$ & \textbf{SDR@2}$\uparrow$ & \textbf{P95}$\downarrow$ \\
\midrule
Zhu et al.~\cite{zhu2021you}              & 67.82\,$\pm$\,18.55 & 132.46\,$\pm$\,30.43 & 9.43\,$\pm$\,3.26  & 370.73\,$\pm$\,57.18 & 37.45\,$\pm$\,11.27 & 39.61\,$\pm$\,10.27 & 19.59\,$\pm$\,4.98  & 266.69\,$\pm$\,92.00 \\
Di Via et al.~\cite{divia2025diffusion}   & 3.44\,$\pm$\,0.57   & 3.01\,$\pm$\,0.67    & 52.00\,$\pm$\,4.12 & 7.92\,$\pm$\,0.80    & 3.10\,$\pm$\,0.59   & 2.71\,$\pm$\,0.41   & 52.93\,$\pm$\,3.35  & 8.12\,$\pm$\,0.79    \\
\midrule
ResNet-101 (ImageNet)                     & 5.18\,$\pm$\,1.24   & 4.07\,$\pm$\,0.43    & 43.87\,$\pm$\,3.76 & 10.23\,$\pm$\,1.56   & 3.09\,$\pm$\,0.27   & 3.16\,$\pm$\,0.34   & 45.67\,$\pm$\,2.21  & 6.91\,$\pm$\,0.17    \\
ResNet-101 (DINO)~\cite{caron2021emerging}  & 4.92\,$\pm$\,0.51   & 5.13\,$\pm$\,0.69    & 43.87\,$\pm$\,2.44 & 15.15\,$\pm$\,4.16   & 4.21\,$\pm$\,0.78   & 4.23\,$\pm$\,1.22   & 44.07\,$\pm$\,2.96  & 12.25\,$\pm$\,4.37   \\
ResNet-101 (MoCo\,v3)~\cite{chen2021mocov3}    & 5.19\,$\pm$\,1.75   & 5.17\,$\pm$\,1.65    & 42.73\,$\pm$\,2.06 & 20.98\,$\pm$\,18.40  & 4.22\,$\pm$\,0.39   & 4.11\,$\pm$\,0.63   & 44.00\,$\pm$\,2.24  & 9.85\,$\pm$\,1.59    \\
ResNet-101 (SimCLR\,v2)~\cite{chen2020big}    & 5.50\,$\pm$\,0.34   & 5.59\,$\pm$\,0.67    & 38.20\,$\pm$\,2.78 & 20.63\,$\pm$\,0.86   & 3.58\,$\pm$\,0.36   & 3.74\,$\pm$\,0.40   & 44.00\,$\pm$\,2.48  & 8.86\,$\pm$\,0.89    \\
\midrule
CDPM Scratch                              & 6.39\,$\pm$\,1.35   & 6.40\,$\pm$\,1.81    & 50.40\,$\pm$\,2.62 & 26.25\,$\pm$\,10.27  & 5.02\,$\pm$\,2.98   & 4.75\,$\pm$\,2.22   & 53.20\,$\pm$\,2.81  & 18.06\,$\pm$\,20.69  \\
CDPM (NIH, $\lambda$=0) Sec. \ref{sec:ablations}                 & \textbf{2.89\,$\pm$\,0.36} & \textbf{2.49\,$\pm$\,0.49} & \textbf{53.40\,$\pm$\,3.18} & \textbf{6.41\,$\pm$\,0.08} & 2.65\,$\pm$\,0.21 & \textbf{2.50\,$\pm$\,0.34} & 53.53\,$\pm$\,3.26 & 6.77\,$\pm$\,0.34 \\
\textbf{CDPM-align (Ours, $\lambda$=5)}   & 3.01\,$\pm$\,0.35   & 2.96\,$\pm$\,0.61    & 53.07\,$\pm$\,2.23 & 7.17\,$\pm$\,0.54    & \textbf{2.58\,$\pm$\,0.28} & 2.58\,$\pm$\,0.36 & \textbf{56.00\,$\pm$\,2.49} & \textbf{6.51\,$\pm$\,0.46} \\
\bottomrule
\end{tabular}}
\end{table}

\begin{table}[t]
\centering
\small
\caption{Landmark detection on \textbf{ISBI2015}. Best results per column are bolded.}
\label{tab:isbi}
\resizebox{\columnwidth}{!}{%
\begin{tabular}{lcccccccc}
\toprule
\multirow{2}{*}{\textbf{Method}}
& \multicolumn{4}{c}{\textbf{10-shot}}
& \multicolumn{4}{c}{\textbf{25-shot}} \\
\cmidrule(lr){2-5} \cmidrule(lr){6-9}
& \textbf{MRE}$\downarrow$ & \textbf{ERE}$\downarrow$ & \textbf{SDR@2}$\uparrow$ & \textbf{P95}$\downarrow$
& \textbf{MRE}$\downarrow$ & \textbf{ERE}$\downarrow$ & \textbf{SDR@2}$\uparrow$ & \textbf{P95}$\downarrow$ \\
\midrule
Zhu et al.~\cite{zhu2021you}              & 17.20\,$\pm$\,4.29  & 48.33\,$\pm$\,9.31   & 38.25\,$\pm$\,3.29 & 91.94\,$\pm$\,11.10  & 2.63\,$\pm$\,0.20   & 3.23\,$\pm$\,0.26   & 66.96\,$\pm$\,2.14  & 5.29\,$\pm$\,0.28    \\
Di Via et al.~\cite{divia2025diffusion}   & 2.70\,$\pm$\,0.34   & 1.97\,$\pm$\,0.19    & 66.57\,$\pm$\,1.39 & 5.41\,$\pm$\,0.47    & 1.80\,$\pm$\,0.10   & 1.26\,$\pm$\,0.10   & 74.06\,$\pm$\,1.74  & 4.34\,$\pm$\,0.20    \\
\midrule
ResNet-101 (ImageNet)                     & 2.83\,$\pm$\,0.30   & 2.38\,$\pm$\,0.37    & 64.20\,$\pm$\,1.19 & 5.42\,$\pm$\,0.68    & 2.86\,$\pm$\,0.43   & 2.37\,$\pm$\,0.38   & 68.83\,$\pm$\,1.28  & 4.87\,$\pm$\,0.36    \\
ResNet-101 (DINO)~\cite{caron2021emerging}  & 4.71\,$\pm$\,0.13   & 3.96\,$\pm$\,0.23    & 54.29\,$\pm$\,1.29 & 18.88\,$\pm$\,1.90   & 2.36\,$\pm$\,0.24   & 1.94\,$\pm$\,0.25   & 66.98\,$\pm$\,1.28  & 5.09\,$\pm$\,0.40    \\
ResNet-101 (MoCo\,v3)~\cite{chen2021mocov3}    & 5.23\,$\pm$\,0.55   & 4.39\,$\pm$\,0.60    & 52.56\,$\pm$\,1.84 & 24.62\,$\pm$\,6.16   & 2.44\,$\pm$\,0.20   & 2.12\,$\pm$\,0.23   & 66.15\,$\pm$\,1.76  & 5.18\,$\pm$\,0.51    \\
ResNet-101 (SimCLR\,v2)~\cite{chen2020big}    & 4.74\,$\pm$\,0.50   & 3.85\,$\pm$\,0.56    & 55.04\,$\pm$\,1.74 & 19.73\,$\pm$\,4.90   & 2.41\,$\pm$\,0.17   & 1.89\,$\pm$\,0.14   & 66.53\,$\pm$\,1.39  & 5.30\,$\pm$\,0.37    \\
\midrule
CDPM Scratch                              & 4.87\,$\pm$\,0.89   & 3.80\,$\pm$\,0.89    & 64.23\,$\pm$\,2.32 & 16.64\,$\pm$\,6.00   & 1.89\,$\pm$\,0.14   & 1.46\,$\pm$\,0.15   & 74.91\,$\pm$\,1.35  & 4.51\,$\pm$\,0.20    \\
CDPM (NIH, $\lambda$=0) Sec. \ref{sec:ablations}                  & 2.32\,$\pm$\,0.15   & 1.78\,$\pm$\,0.27    & \textbf{68.12\,$\pm$\,1.40} & 5.20\,$\pm$\,0.20 & 1.70\,$\pm$\,0.06   & 1.19\,$\pm$\,0.04   & 75.49\,$\pm$\,1.43  & 4.04\,$\pm$\,0.15    \\
\textbf{CDPM-align (Ours, $\lambda$=5)}   & \textbf{2.11\,$\pm$\,0.29} & \textbf{1.38\,$\pm$\,0.14} & 67.48\,$\pm$\,3.20 & \textbf{4.75\,$\pm$\,0.37} & \textbf{1.54\,$\pm$\,0.02} & \textbf{0.95\,$\pm$\,0.07} & \textbf{77.52\,$\pm$\,0.74} & \textbf{3.90\,$\pm$\,0.07} \\
\bottomrule
\end{tabular}}
\end{table}

\begin{table}[t]
\centering
\small
\caption{Landmark detection on \textbf{DHA}. Best results per column are bolded.}
\label{tab:dha}
\resizebox{\columnwidth}{!}{%
\begin{tabular}{lcccccccc}
\toprule
\multirow{2}{*}{\textbf{Method}}
& \multicolumn{4}{c}{\textbf{10-shot}}
& \multicolumn{4}{c}{\textbf{25-shot}} \\
\cmidrule(lr){2-5} \cmidrule(lr){6-9}
& \textbf{MRE}$\downarrow$ & \textbf{ERE}$\downarrow$ & \textbf{SDR@2}$\uparrow$ & \textbf{P95}$\downarrow$
& \textbf{MRE}$\downarrow$ & \textbf{ERE}$\downarrow$ & \textbf{SDR@2}$\uparrow$ & \textbf{P95}$\downarrow$ \\
\midrule
Zhu et al.~\cite{zhu2021you}              & 33.85\,$\pm$\,13.68 & 90.66\,$\pm$\,9.75   & 44.48\,$\pm$\,6.51  & 161.83\,$\pm$\,33.82 & 2.87\,$\pm$\,0.43   & 4.95\,$\pm$\,0.89   & 83.13\,$\pm$\,0.98  & 4.04\,$\pm$\,0.40    \\
Di Via et al.~\cite{divia2025diffusion}   & 4.34\,$\pm$\,1.04   & 3.01\,$\pm$\,0.81    & 79.94\,$\pm$\,3.25  & 19.11\,$\pm$\,11.34  & 2.22\,$\pm$\,0.15   & 1.48\,$\pm$\,0.09   & 88.28\,$\pm$\,0.41  & 3.67\,$\pm$\,0.10    \\
\midrule
ResNet-101 (ImageNet)                     & 6.07\,$\pm$\,0.64   & 5.76\,$\pm$\,0.30    & 78.44\,$\pm$\,0.36  & 26.39\,$\pm$\,3.29   & 3.95\,$\pm$\,1.12   & 4.39\,$\pm$\,1.26   & 84.07\,$\pm$\,1.48  & 7.65\,$\pm$\,4.80    \\
ResNet-101 (DINO)~\cite{caron2021emerging}  & 23.63\,$\pm$\,6.96  & 22.38\,$\pm$\,5.80   & 34.95\,$\pm$\,8.35  & 99.49\,$\pm$\,25.50  & 10.16\,$\pm$\,6.97  & 10.24\,$\pm$\,6.34  & 63.83\,$\pm$\,15.43 & 46.98\,$\pm$\,23.90  \\
ResNet-101 (MoCo\,v3)~\cite{chen2021mocov3}    & 24.60\,$\pm$\,8.45  & 23.48\,$\pm$\,7.37   & 34.86\,$\pm$\,11.53 & 102.67\,$\pm$\,24.80 & 9.28\,$\pm$\,4.21   & 9.64\,$\pm$\,4.32   & 64.50\,$\pm$\,10.54 & 45.29\,$\pm$\,17.23  \\
ResNet-101 (SimCLR\,v2)~\cite{chen2020big}    & 20.24\,$\pm$\,3.22  & 19.91\,$\pm$\,3.19   & 39.36\,$\pm$\,4.77  & 86.83\,$\pm$\,8.75   & 9.47\,$\pm$\,4.37   & 9.82\,$\pm$\,4.56   & 64.97\,$\pm$\,10.17 & 46.99\,$\pm$\,18.75  \\
\midrule
CDPM Scratch                              & 8.60\,$\pm$\,2.57   & 5.83\,$\pm$\,2.06    & 71.19\,$\pm$\,4.57  & 44.34\,$\pm$\,13.61  & 4.22\,$\pm$\,0.85   & 2.96\,$\pm$\,0.69   & 84.50\,$\pm$\,1.56  & 14.43\,$\pm$\,9.88   \\
CDPM (NIH, $\lambda$=0) Sec. \ref{sec:ablations}                 & 4.08\,$\pm$\,0.75   & 3.24\,$\pm$\,0.57    & 80.50\,$\pm$\,2.20  & 16.41\,$\pm$\,4.22   & 1.62\,$\pm$\,0.11   & 1.19\,$\pm$\,0.14   & 89.80\,$\pm$\,0.42  & 3.18\,$\pm$\,0.16    \\
\textbf{CDPM-align (Ours, $\lambda$=5)}   & \textbf{2.51\,$\pm$\,0.30}   & \textbf{2.16\,$\pm$\,0.31}    & \textbf{86.47\,$\pm$\,1.08}  & \textbf{4.46\,$\pm$\,0.90}    & \textbf{1.53\,$\pm$\,0.10} & \textbf{0.97\,$\pm$\,0.08} & \textbf{91.13\,$\pm$\,0.69} & 2.77\,$\pm$\,0.20 \\
\bottomrule
\end{tabular}}
\end{table}

\paragraph{Reliability.}
At 25-shot, CDPM-align achieves ERE below 1\,mm on both ISBI2015 (0.95\,mm, a 25\% improvement over~\cite{divia2025diffusion} reaching 1.26\,mm) and DHA (0.97\,mm, a 34\% improvement over~\cite{divia2025diffusion}  reaching 1.48\,mm). Consistently, ERE $\leq$ MRE across conditions, indicating tight probability mass concentration. Figure~\ref{fig:ere_comparison} provides visual corroboration: CDPM-align systematically produces tight, unimodal heatmaps in contrast to the diffuse distributions of other methods. P95 improvements track this pattern across all benchmarks.

\paragraph{Robustness.}
SSL methods using discriminative objectives exhibit severe performance degradation on DHA at 10-shot: DINO, MoCo\,v3, and SimCLR\,v2 reach MRE of 20--25\,mm with P95 exceeding 86\,mm, compared to 4.46\,mm for CDPM-align. This failure is consistent across all 5 independent runs and persists at 25-shot (9--10\,mm MRE). When pooled multi-anatomical regions X-rays span heterogeneous structural modes, discriminative SSL objectives collapse onto global image identity, suppressing fine-grained intra-region geometry. Both CDPM-align and Di~Via et~al.\ maintain stable performance, with CDPM-align providing additional gains from the alignment objective.

\subsection{Ablation Studies}
\label{sec:ablations}

\paragraph{Alignment weight $\lambda$.}
We analyse the effect of the alignment weight $\lambda$ on Shenzhen (Table~\ref{tab:ablation}), the smallest and structurally simplest benchmark, to isolate dataset-independent trends. 
Without alignment ($\lambda{=}0$), MRE is 3.95\,px at 10-shot. Increasing $\lambda$ improves MRE and SDR up to $\lambda{=}5$ (best MRE: 3.01\,px; SDR: 53.07\%; P95 at 25-shot: 6.51\,px). Beyond this, performance drops ($\lambda{=}10$: 3.42\,px MRE), indicating that over-constraining the guidance reduces pixel-level fidelity. 
This trade-off reflects the differing roles of features along the diffusion trajectory: larger $\lambda$ forces low-noise features to rigidly follow coarse global semantics, over-concentrating probability mass and minimising ERE, but limiting the flexibility required for precise sub-millimetre localisation. ERE is lowest at $\lambda{=}10$, highlighting that minimal uncertainty does not necessarily translate into optimal spatial accuracy. Based on these observations, we adopt $\lambda{=}5$ as the operating point that best balances accuracy, uncertainty, and robustness.

\paragraph{Pretraining data scale.}
To further assess the benefit of the proposed feature alignment, we design an ablation study comparing CDPM-align against regular CDPM pre-trained with a larger dataset. As such, we train the same CDPM architecture, with $\lambda{=}0$, on NIH ChestX-ray14~\cite{wang2017chestxnet} (${\sim}$112k chest X-rays), referred to as CDPM (NIH). Results in Tables~\ref{tab:shenzhen}--\ref{tab:dha} show that CDPM-align (988 images) is competitive with CDPM (NIH) on Shenzhen 25-shot (2.58 vs.\ 2.65\,px MRE), where CPDM (NIH) benefits from in-domain images, while providing slight improvements on the out-of-domain DHA 10-shot (2.51 vs.\ 4.08\,mm) and ISBI2015 10-shot (2.11 vs.\ 2.32\,mm), where NIH domain shift penalises the larger corpus. The alignment objective compensates for reduced data scale by producing more discriminative representations from limited heterogeneous data.

\begin{table}[t]
\centering
\small
\caption{Ablation of alignment weight $\lambda$ on \textbf{Shenzhen}.}
\label{tab:ablation}
\resizebox{0.8\columnwidth}{!}{%
\begin{tabular}{lcccccccc}
\toprule
\multirow{2}{*}{\textbf{$\lambda$}}
& \multicolumn{4}{c}{\textbf{10-shot}}
& \multicolumn{4}{c}{\textbf{25-shot}} \\
\cmidrule(lr){2-5} \cmidrule(lr){6-9}
& \textbf{MRE}$\downarrow$ & \textbf{ERE}$\downarrow$ & \textbf{SDR@2}$\uparrow$ & \textbf{P95}$\downarrow$
& \textbf{MRE}$\downarrow$ & \textbf{ERE}$\downarrow$ & \textbf{SDR@2}$\uparrow$ & \textbf{P95}$\downarrow$ \\
\midrule
0  & 3.95\,$\pm$\,0.84 & 3.35\,$\pm$\,1.12 & 51.60\,$\pm$\,3.52 & 9.30\,$\pm$\,1.56 & 2.75\,$\pm$\,0.30 & 2.69\,$\pm$\,0.32 & 53.20\,$\pm$\,3.68 & 7.20\,$\pm$\,0.44 \\
1  & 3.82\,$\pm$\,1.07 & 3.28\,$\pm$\,1.06 & 51.53\,$\pm$\,2.69 & 9.44\,$\pm$\,3.98 & 2.61\,$\pm$\,0.17 & 2.59\,$\pm$\,0.28 & 53.47\,$\pm$\,3.20 & 7.00\,$\pm$\,0.69 \\
3  & 3.25\,$\pm$\,0.86 & 2.88\,$\pm$\,0.56 & 52.20\,$\pm$\,0.80 & 7.03\,$\pm$\,0.39 & 2.89\,$\pm$\,0.54 & 2.57\,$\pm$\,0.39 & 53.47\,$\pm$\,1.56 & 7.26\,$\pm$\,0.46 \\
5  & 3.01\,$\pm$\,0.35 & 2.96\,$\pm$\,0.61 & 53.07\,$\pm$\,2.23 & 7.17\,$\pm$\,0.54 & 2.58\,$\pm$\,0.28 & 2.58\,$\pm$\,0.36 & 56.00\,$\pm$\,2.49 & 6.51\,$\pm$\,0.46 \\
10 & 3.42\,$\pm$\,0.46 & 2.78\,$\pm$\,0.22 & 52.67\,$\pm$\,1.75 & 7.23\,$\pm$\,0.44 & 2.74\,$\pm$\,0.32 & 2.37\,$\pm$\,0.10 & 53.93\,$\pm$\,0.80 & 7.23\,$\pm$\,0.56 \\
\bottomrule
\end{tabular}}
\end{table}
\section{Conclusion}
\label{sec:conclusion}
We presented CDPM-align, a conditional diffusion pretraining framework with multi-scale guidance alignment for anatomical landmark detection under realistic low-annotation clinical budgets. By enforcing directional consistency of the class-conditional guidance signal $\Delta h$ across diffusion timesteps and UNet levels, the model learns stable, dataset-specific representations from pooled heterogeneous X-rays.
Across Shenzhen, ISBI2015, and DHA in 10- and 25-shot regimes, CDPM-align consistently improves accuracy (MRE), detection rate (SDR), tail reliability (P95), and uncertainty concentration (ERE) over supervised, self-supervised, unconditional diffusion, and large-scale domain-specific pretraining. In contrast to conventional SSL, which degrades on heterogeneous datasets, CDPM-align remains robust across domains. Notably, pretrained on only 988 images, it matches or surpasses models trained on 112k NIH radiographs, with the largest gains on DHA where domain shift penalises out-of-domain corpora. Sub-millimetre ERE at 25-shot on ISBI2015 and DHA further demonstrates reliable localisation with minimal supervision.
Overall, targeted in-domain conditional diffusion pretraining emerges as a practical and deployment-oriented alternative to large out-of-domain training for clinical imaging. Current limitations include evaluation restricted to 2D X-rays and a moderated increase in  computational cost during alignment ($1.3 \times$ with respect to naive CDPM training), pointing to future work on scaling to volumetric modalities, while further improving efficiency.\\
{\bf Disclosure of interests.} The authors have no competing interests to declare.

%
%
%
\bibliographystyle{splncs04}
\bibliography{references_short}

\end{document}